\documentclass[11pt]{article}

\makeatletter

\let\proof\@undefined
\let\endproof\@undefined
\makeatother

\usepackage[utf8]{inputenc}
\usepackage[ruled, linesnumbered]{algorithm2e}
\usepackage{graphicx}
\usepackage{xcolor}
\usepackage{amsthm}
\usepackage{amsfonts}
\usepackage{hyperref}
\usepackage{multirow}
\usepackage{lineno}
\usepackage{amsmath}
\usepackage[normalem]{ulem}
\usepackage{diagbox}
\usepackage{tabularx}
\usepackage{subfig}
\usepackage{float}
\usepackage[a4paper, margin=1in]{geometry}

\usepackage[natbibapa]{apacite}
\usepackage[english]{babel}

\usepackage{collectbox}

\makeatletter
\newcommand{\sqbox}{%
    \collectbox{%
        \@tempdima=\dimexpr\width-\totalheight\relax
        \ifdim\@tempdima<\z@
            \fbox{\hbox{\hspace{-.1\@tempdima}\BOXCONTENT\hspace{-.1\@tempdima}}}%
        \else
            \ht\collectedbox=\dimexpr\ht\collectedbox+.2\@tempdima\relax
            \dp\collectedbox=\dimexpr\dp\collectedbox+.0\@tempdima\relax
            \fbox{\BOXCONTENT}%
        \fi
    }%
}
\makeatother

\newtheorem{definition}{Definition}
\newtheorem{theorem}{Theorem}
\newtheorem{lemma}{Lemma}

\newtheorem{remark}{Remark}

\newcommand{\takeoffline}[0]{\text{$r_{top}$}}
\newcommand{\pLvisible}[0]{$p_L$-\textit{visible}}
\newcommand{\nonpLvisible}[0]{\textit{non $p_L$-visible}}
\newcommand{\cLvisible}[0]{$c_L$-\textit{visible}}
\newcommand{\noncLvisible}[0]{\textit{non $c_L$-visible}}
\newcommand{\approach}[0]{MASPA}

\title{MASPA: An efficient strategy for path planning with a tethered marsupial robotics system}

\renewcommand\footnotemark{}

\author{Jes\'us Capit\'an$^{1}$, Jos\'e M. D\'iaz-B\'añez$^{2}$, Miguel A. P\'erez-Cutiño$^{2}$, \\ Fabio Rodr\'iguez$^{2,*}$ and Inmaculada Ventura$^{2}$
\thanks{$^{1}$ J. Capit\'an is with the Multi-robot \& Control Systems group, University of Seville, Spain,
        {\tt\small {jcapitan}@us.es}}%
\thanks{$^{2}$ J.M. D\'iaz-B\'añez, M.A. P\'erez-Cutiño, F. Rodr\'iguez and I. Ventura are with the Department of Applied Mathematics II, University of Seville, Spain,
        {\tt\small [dbanez,mpcutino,frodriguex,iventura]@us.es}}%
\thanks{$*$ Corresponding Author}%
 }

\begin{document}

\maketitle

\begin{abstract}
A tethered marsupial robotics system comprises three components: an Unmanned Ground Vehicle (UGV), an Unmanned Aerial Vehicle (UAV), and a tether connecting both robots. Marsupial systems are highly beneficial in industry as they extend the UAV's battery life during flight. This paper introduces a novel strategy for a specific path planning problem in marsupial systems, where each of the three components must avoid collisions with ground and aerial obstacles modeled as 3D cuboids. Given an initial configuration in which the UAV is positioned atop the UGV, the goal is to reach an aerial target with the UAV. We assume that the UGV first moves to a position from which the UAV can take off and fly through a vertical plane to reach an aerial target. We propose an approach that discretizes the space to approximate an optimal solution, minimizing the sum of the lengths of the ground and air paths. First, we assume a taut tether and use a novel algorithm that leverages the convexity of the tether and the geometry of obstacles to efficiently determine the locus of feasible take-off points for the UAV. We then apply this result to scenarios that involve loose tethers. 
The efficiency of our method enables real-time decision-making, making it suitable for use in emergency situations where quick responses are crucial. The simulation test results show that our approach can solve complex situations in seconds, outperforming a baseline planning algorithm based on RRT* (Rapidly exploring Random Trees). 
\end{abstract}

\begin{center}
\end{center}

\section{Introduction}


Mobile robots are intelligent devices capable of autonomously performing specific tasks in complex environments, thereby reducing the risks and dangers that these environments may pose to humans.
Particulary, designing and managing drone-based systems presents significant challenges from both a modeling and problem-solving perspective. Addressing these complexities requires the integration of optimization techniques
with efficient computational methods to ensure effective and reliable implementation of these systems \citep{chung2020optimization}.
A specific area within this field is the path planning in 
combined robotics systems, which has become increasingly important in addressing challenges in various applications, particularly in confined or hazardous environments. In these systems, both aerial and ground-based robots can work together, offering a diverse set of capabilities for effective path planning and trajectory optimization. Aerial robots offer superior maneuverability, while ground-based robots provide greater payload capacity and longer operational autonomy. 
In this context, research efforts are centered on identifying optimal paths for applications such as emergency missions, including firefighting, search and rescue, and surveillance, where efficient path planning is crucial \citep{MADRIDANO2021114660}.

In robotics, path planning can generally be divided into global and local path planning, depending on the level of information available about the environment \citep{liu2023path}. Global path planning assumes that the robot has complete knowledge of the environment and can follow a predefined path to reach its target. Due to this characteristic, global path planning is often referred to as offline or static path planning. In contrast, local path planning assumes that the robot has partial or no prior knowledge of the environment, requiring real-time monitoring and adaptive responses. This approach is known as online or dynamic path planning because it adjusts the robot’s path based on immediate environmental data.

In aerial robotics, the limited autonomy of drones is a significant disadvantage in tasks that demand quick response times.
Then, the integration of heterogeneous systems, combining Unmanned Aerial Vehicles (UAVs) and Unmanned Ground Vehicles (UGVs), further enhances the efficiency of these operations \citep{xia2023two}.

In this work, we focus on marsupial robotics systems~\citep{hourani_chapter13}, which are teams of robots with heterogeneous capabilities, where some act as providers to others. Typically, the provider deploys other robots and serves as a communication relay and/or recharge station. We are interested in tethered marsupial systems; in particular, a system operating in a confined space where a UGV deploys a tethered UAV to extend the autonomy of the latter through the cable.
Planning 3D collision-free paths for the operation of such tethered marsupial systems in confined scenarios is challenging in several manners. First, due to the high-dimensionality of the configuration space, as both robots present different locomotion capabilities and their motion needs to be planned in a coordinated manner. Second, the fact that the movement of both robots is coupled by a tether imposes additional constraints. Not only does the length of the tether have to be considered when planning paths, but also its shape to ensure that there are no cable entanglements or collisions, particularly in cluttered environments. 

In this paper, we propose a method for 3D collision-free path planning of a marsupial system composed of a UGV that deploys a tethered UAV. We follow a sequential strategy for robot navigation: given a starting point of the marsupial system and a destination point to be visited by the UAV, we assume that the UGV will move first carrying the UAV to a location from where the final destination is accessible by the tethered UAV. The UAV is allowed to fly in a 2D vertical plane from above the UGV to the target. 
To address this, we introduce the \emph{Catenary Visibility Problem}, in which we identify candidate locations from which the UGV can safely deploy the UAV without the risk of collision.
This problem is novel and interesting in its own right.
We then compute a 2D path for the UAV starting at the candidate point that minimizes the time to reach the aerial target. We call our strategy \approach{}, which stands for {MA}rsupial {S}equential {P}ath-planning {A}pproach.

We assume that the tether is controllable and bounded by a maximum length $L$. We first study the case where the tether is taut and forms a polygonal chain, simulating a cable that is tangent to an arbitrary number of obstacles. In that case, we take advantage of the geometric properties of the problem and propose an efficient $O(n^2)$ time algorithm, where $n$ is the total number of vertices in the obstacles, to find the exact locus of take-off points in a vertical plane from where the UAV can be deployed to reach the target (2D catenary visibility).  
Then, we extend the algorithm to approximate the locus of feasible take-off points in the space (3D catenary visibility). When approaching the catenary-visibility problem in 3D, we perform a radial division of the space in a set of vertical planes crossing the destination point. We name the 3D version of the algorithm as PVA, after \textbf{P}olygonal \textbf{V}isibility \textbf{A}lgorithm, and use it as a key component in the \approach{} strategy. In addition, we consider the case where the loose tether is realistically modeled as a catenary curve. It is important to note that PVA enables the construction of a linear structure, eliminating the need for time-consuming queries regarding the existence of collision-free catenaries, which is a major bottleneck in current algorithms \citep{martinez2023path}; hence, PVA can significantly reduce the search time for viable loose tethers, which is crucial in emergency scenarios.

\subsection{Related Work}

The Multi-Robot Path Planning (MRPP) problem consists of computing collision-free paths for multiple robots from start locations to goal locations, minimizing different objectives based on covered distance or arrival time. 
Zafar et al. \citep{ZAFAR2018141} classified mobile robot path planning methods into four categories: Analytical, Enumerative, Evolutionary, and Meta-Heuristic. In particular, analytical approaches leverage the geometric properties of the problem to find theoretically optimal solutions. 

Numerous methods for robot path planning assume that the environment can be modeled by a graph and then compute optimal paths using graph search algorithms like Djikstra, A$^\star$, or Theta*~\citep{nash_aaai10}. Specially, the A$^\star$ algorithm has been used for path planning and optimization problems by many authors in the operational research field 
\citep{yuan2016research, kiadi2023based}. 
Probabilistic planners~\citep{elbanhawi_access14}, such as Probabilistic Road-Maps (PRM) or Rapidly-exploring Random Trees (RRT), are also a widespread approach. 
Instead of considering a fully connected scenario, they build a graph by randomly sampling the environment, which makes them suitable for planning in high-dimensional spaces in reasonable computation time. Some variants, such as RRT$^\star$~\citep{karaman_ijrr11} or multi-RRT \citep{karaman_ijrr11, feng2024adaptive}, can achieve optimal solutions asymptotically. 
Recently, reinforcement learning methods have also been applied to UAV navigation in complex environments~\citep{wang_tvt19}, even considering multi-agent teams~\citep{liu2020mapper}. 
Another constrained optimization alternative for MRPP is Non-linear Model Predictive Control, where a receding time horizon approach is used to compute collision-free trajectories for multiple robots~\citep{zhu_ral19}.

More specifically, this work focuses on path planning for marsupial multi-robot systems that combine ground and aerial robots connected by a tether~\citep{murphy1999marsupial}. Dinelli et al.~\citep{dinelli_drones23} wrote a review of UGV-UAV robotic systems for operation in underground rescue missions. Hierarchical trajectory generation has been proposed for marsupial robot systems~\citep{stan_iros19}, combining high-level multi-robot path planning on a topological graph that encodes the locomotion capabilities of each robot, with low-level dynamically feasible trajectory planning through RRT and non-linear programming. Although they plan trajectories that take into account the whole marsupial team in a coordinated manner, no specific constraints related to tethered systems are considered. 

Sandino et al. \citep{sandino2014tether} discuss the problem of landing a helicopter when GPS sensors are not reliable. In this case, using a tether connecting the helicopter to the base allows for the estimation of its linear position relative to the landing point. Moreover, the tension exerted on the tether provides a stabilizing effect on the helicopter's translational dynamics. However, attached UAVs impose additional complexities in addressing cable disturbances in the robot controller~\citep{tognon_book20} and avoiding cable entanglement~\citep{cao_tro23}. Viegas et al.~\citep{viegas_jint22} presented a novel lightweight tethered UAV with mixed multi-rotor and water jet propulsion for forest fire fighting. The planning of tether-aware kinodynamic trajectories in formations with multiple tethered UAVs has also been considered~\citep{cao_tro23,bolognini_cdc22}. Nevertheless, these works assume that ground stations are static. 
There are also works on tethered UAV-UGV marsupial systems, with moving ground stations. A sensor system that measures the catenary shape of the cable can be used for relative localization~\citep{borgese_ral22}. Miki et al.~\citep{miki_icra19} presented a cooperative tethered UAV-UGV system where the UAV can anchor the tether on top of a cliff to help the UGV climb, using a grid-based heuristic planner (A$^\star$) for navigation. Furthermore, a hierarchical path planning approach with two independent RRT$^\star$ has been proposed~\citep{papa_med14} for map generation missions. The UGV first plans its route and then the UAV limits its range according to the UGV plan and the tether length. Both methods approach motion planning independently for each robot, neglecting collision avoidance for the tether. 

In the work by Xiao et al.~\citep{xiao_ssrr19}, for a given position of the UGV, the UAV follows the path using motion controllers that consider the relative angles and length of the tether. The same authors extended their previous work to build a marsupial robot system for search\&rescue operation~\citep{xiao2021autonomous}, with a teleoperated ground robot and an autonomous tethered UAV that provides visual feedback and situational awareness to the teleoperator. They implemented a probabilistic model for risk-aware path planning and even planned contact points of the tether with the environment to extend the UAV line of sight.

In a setting more closely related to ours,
Mart\'inez-Rozas et al.~\citep{martinez-rozas_icra21} applied RRT$^*$ and non-linear optimization based on sparse factored graphs to compute collision-free trajectories for a marsupial system with a UGV, a UAV, and a non-taut tether with controllable length. They consider a realistic model of the tether shape for collision checking, but the UGV is assumed to be static for trajectory planning.  Later, the same authors propose a method for designing a collision-free path for both the UAV and the UGV, considering the complexities of the 3D environment and the constraints imposed by the tether (\cite{martinez2023path}). 
A coordinated parallel movement of the two vehicles, linked by a cable modeled using a catenary curve, is proposed. However, incorporating the catenary equation into the process makes the planning task computationally expensive, resulting in the need for the planning process to be conducted offline.

In this paper, we propose an efficient sequential planning strategy in which the UGV advances to a specific point, from which the UAV takes off to reach the desired target. This setup, where the UGV carries the UAV, was justified by Nicolas Hudson et al.~\citep{hudson2021heterogeneous}, emphasizing its potential in scenarios requiring coordinated robotic operations. 
The same setup has also been used for the inspection of underground stone mine pillars \citep{martinez2023oxpecker}. The UAV stays landed on the UGV while the ensemble moves inside a mine. The mission of the UAV is to create 3D maps of the mine pillars to support time-lapse hazard mapping and time-dependent pillar degradation analysis.

\subsection{Our Contribution}

Overall, optimal methods for MRPP suffer from scalability, and many try to alleviate the complexity of the problem by proposing hierarchical approaches or decoupled planning for robots. In our tethered marsupial system operating in 3D, due to the coupled nature of the motion, path planning needs to be performed in a high-dimensional configuration space that takes into account both robots. Therefore, there is a need for an efficient method capable of jointly computing optimal robot paths in a reasonable time. Furthermore, most existing MRPP methods do not consider the specific constraints imposed by a tether for collision avoidance. 

To the best of our knowledge, there are no path planning approaches equivalent to \approach{} for tethered UAV-UGV marsupial systems in 3D confined environments. Even though we assume that the movements of the UGV and the UAV are sequential, we tackle the complexity of the path planning problem in a high-dimensional search space, as we pursue optimal motion for both robots. Also, we carry out collision checking by integrating a realistic model for the tether geometry. 
The visibility challenges introduced by the catenary present a novel problem in the field of computational geometry, requiring new approaches to efficiently determine collision-free paths while considering the complex geometry of tethered systems.

This paper offers multiple contributions, which are organized as follows:  
Section \ref{sec:model} outlines the model of the marsupial system we are employing. In Section \ref{sec:problem}, we introduce the optimization path planning problem. Section \ref{sec:visibility} presents an efficient algorithm to solve the visibility problem using a taut tether. This approach enables an efficient resolution of the path planning problem, which is addressed in Section \ref{sec:opt}. Then, we further elaborate on the applicability of our algorithm in the case where the controllable tether is modeled as a catenary curve, Section \ref{sec:extension}. We conducted a study on the parameters of our planner \approach{} in Section \ref{sec:parameter_study}, 
where its performance was evaluated across random scenarios. Additionally, Section \ref{sec:experiments} focuses on evaluating and comparing our strategy with a baseline algorithm in realistic scenarios.
Finally, conclusions and a number of problems for further research are set out in Section \ref{sec:conclusions}.

\section{The Model}\label{sec:model}

\begin{figure}
\centering
\includegraphics[width=0.6\textwidth]{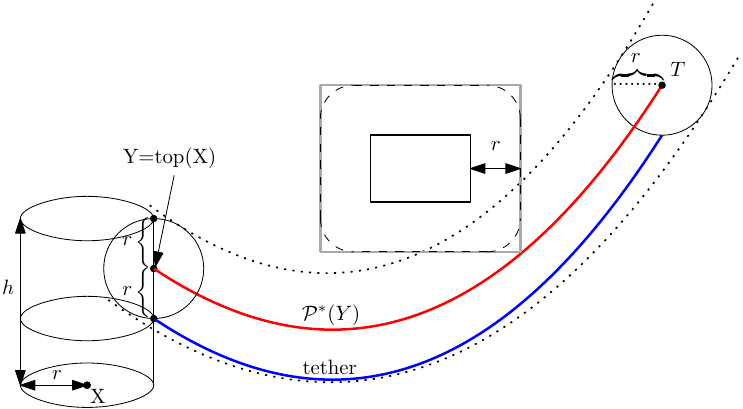}
\caption{Model of the marsupial system with the UGV and the UAV positioned at $X$ and $T$, respectively. The UAV path with minimum length between $Y=top(X)$ and $T$ is denoted as $\mathcal{P}^*(Y)$ and has the same shape as the tether. An enlarged obstacle is also depicted.}
\label{fig:marsupial-system2}
\end{figure}

A tethered marsupial robotics system comprises three elements: a UGV, a UAV, and a controllable tether connecting the two vehicles. 
As mentioned above, in our model, the movement of these vehicles is sequential; that is, the UGV moves when the UAV is stationary on board, and conversely, the UAV flies when the UGV is stopped at a fixed position. When the UGV carries the UAV, we represent the entire system as a cylinder with radius $r$ and height $h$, where $h>2r$; when flying, the UAV is modeled as a sphere of radius $r$. Notice that we consider similar radii for UGV and UAV for the sake of simplicity, but our methods are also valid for cases where they have different sizes. We define a tie point for each vehicle, which is the location where it connects to the tether. For the UAV, the tie point is assumed to be at the bottom of the sphere. For the UGV, the tether is attached to a point on the cylinder with a height of $h-2r$. Figure~\ref{fig:marsupial-system2} illustrates the complete model of the marsupial system.

We assume that the reference position $X$ of the UGV is located at the lowest point of the cylinder's axis. Given the UGV with fixed position $X$ and the aerial target $T$, we establish the UAV take-off point $top(X)$ as a fixed point on the cylinder at height $h-r$. While the UAV is landed, the UGV can rotate to align $top(X)$ within the vertical plane $\pi_{X,T}$ that contains $X$ and $T$ (its cylindrical model allows the system to rotate without colliding if it was initially free of collisions). Before taking off, we despise the movement of the UAV from its landed position on board the UGV to $top(X)$ and assume that there is no collision. When flying, we restrict the UAV motion to the vertical plane $\pi_{X,T}$ following a trajectory from $top(X)$ to $T$. 

In this work, we model obstacles as orthogonal cuboids~\footnote{An orthogonal cuboid is a solid whose edges are all aligned with pairs of orthogonal coordinate axes.}.
This is not a strong limitation from a practical point of view, as typical objects (e.g., walls, beams, or boxes) can be modeled as cuboids, and more complex structures can also be constructed by combining orthogonal cuboids. 

For handling collisions involving both the UAV and the tether, we consider the following definition. We refer the UAV's path to the path followed by the center of the sphere, and we say that \emph{the UAV's path is collision-free} if the locus of points equidistant from the path at distance $r$ does not cross any obstacle. With this definition, a tether below the UAV's path at vertical distance $r$ does not intersect any obstacle. See Figure \ref{fig:marsupial-system2}. 
We assume that the UAV has a path tracker that allows it to fly paths with the same shape as the tether.

The following transformation is useful for collision checking.  
We expand each obstacle by performing the Minkowski sum with the sphere of radius $r$ and consider the minimum cuboid that contains the expanded obstacle. Then, we shrink the UAV sphere down to its central point and we say that the UAV's path is collision-free when it does not cross any of the enlarged cuboids. With this transformation, the geometry of the obstacles is preserved, avoiding curved boundaries that complicate computation and are impractical in real applications. 
Since we consider a taut tether, we assume that no collision occurs if the path is tangent to an obstacle.  
Additionally, with this transformation, 
the cylinder representing the marsupial robotics system (while carrying the UAV)
shrinks down to a vertical segment of length $h-r$ and we say that \emph{the UGV's path is collision-free} if this vertical segment does not cross any of the enlarged obstacles.

\section{Optimization Problem and Strategy Overview}\label{sec:problem}

Given the marsupial robotics system described in Section~\ref{sec:model}, we address an optimization path planning problem where the goal is to minimize the time to reach a given aerial target point. We assume a strategy where the UGV first navigates (with the UAV onboard) to a ground point from which the UAV can take off and reach the target. We consider a constant traveling speed for both vehicles, so the objective function to optimize is the sum of the distances traveled by the vehicles, that is, the sum of the lengths of the UGV path and the UAV path. We formally define the optimization problem as follows:


\textbf{Shortest Marsupial Path Problem (SMPP):} \emph{Let $h$, $r$ and $L$ be positive numbers. Given a 3D scenario with a set $\mathcal{O}$ of $n$ orthogonal cuboids, a point $S$ in the plane $z=0$ (the ground), and a target point $T$ in the aerial region $z>h$, we want to find a point $X$ on the ground such that:}

\begin{enumerate}
    \item \emph{There exists a path $P_g$ from $S$ to $X$ so that the marsupial robot (modeled as a vertical segment of height $h-r$) can traverse $P_g$ without colliding with obstacles in $\mathcal{O}$.}
    
    \item \emph{Given the vertical plane $\pi_{X,T}$ passing through $X$ and $T$, there exists a collision-free path $P_a$ from $top(X)$ to $T$ within $\pi_{X,T}$ for the UAV, , which is modeled as a point. 
    }
    
    \item \emph{There exists a controllable, loose tether of length at most $L$ that can follow, without collision, the UAV along the path $P_a$. }
    
     \item \emph{The sum of the lengths of $P_g$ and $P_a$ is minimized. }
\end{enumerate}

\vspace{.25cm}

The tether can be modeled either with a catenary curve (loose tether) or with a polygonal chain (taut tether). Firstly, we approximate the solution of the SMPP
by modeling the taut tether as an increasing convex polygonal chain, which will be defined later. This chain is generally supported by the obstacles encountered along the path.
Subsequently, we demonstrate that this approach can significantly enhance computational efficiency in problems involving catenaries. 

In the following, we assume a 3D scenario with a set $\mathcal{O}$ of $n$ orthogonal cuboids and a marsupial system model as defined above. 

\begin{definition}
    Given $L > 0$, 
    we say that the point $top(X)$ is $L$-polygonal-visible  (\pLvisible{}, for short) if there exists a collision-free, taut tether, modeled as a polygonal chain, connecting $top(X)$ and $T$ with a length of at most $L$.
\end{definition}

We introduce the following two subproblems that contribute to the solution of the SMPP:

\vspace{.25cm}

\textbf{Polygonal Visibility Problem (PVP):}
\emph{Given two positive numbers $h$ and $r$, with $h>2r$, compute the locus of \pLvisible{} points within the horizontal plane $\pi_{top} \equiv  z=h-r$.}

\vspace{.25cm}

\textbf{Minimum Length 
Tether Problem (MLTP):} \emph{Given $top(X)$ and $T$,
compute the shortest length of a collision-free taut tether, if it exists, connecting $top(X)$ and $T$, and contained in the vertical plane $\pi_{X,T}$.}

\vspace{.25cm}

\begin{figure}[t]
    \centering
    \includegraphics[width=0.6\textwidth, page=1]{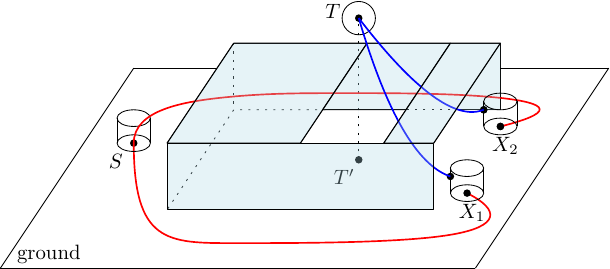}
    \caption{An scenario of the marsupial path planning problem. Two paths are required, a ground path (red) for the UGV carrying the UAV from the starting point $S$ to a ground point $X$ from which the UAV can reach the target $T$ using a collision-free aerial path (blue). In the scenario, two possible solutions are showed; the  general problem is to find the one that minimizes the sum of the lengths.
    }
    \label{fig:planning}
\end{figure}

Assuming a taut tether, the 
strategy proposed in this work to solve the SMPP, referred to as \approach{} (Marsupial Sequential Path-Planning Approach), is based on the following key steps:
we first compute
a discrete set of \pLvisible{} candidate take-off points in the space, each one with an associated UGV candidate position on the ground. Then, we create a visibility graph generated by the initial point $S$, the ground obstacles, and the candidate positions on the ground. Finally, we use this graph to plan a collision-free path for the UGV from $S$ to the best candidate point from which the UAV can take off and reach $T$, so that the sum of the aerial and ground paths is minimized. 
In addition, we extend these ideas to apply to scenarios that involve a loose tether instead of a taut one. Figure \ref{fig:planning} shows a scenario of the marsupial path planning problem and two possible solutions.

\section{Polygonal Visibility Problem}\label{sec:visibility}

In this Section, we propose a competitive algorithm to tackle the Polygonal Visibility Problem (PVP). To the best of our knowledge, this geometric problem has not been previously addressed in the literature.
First, we define a 2D version of the problem (PVP-2D), where we search for the set of \pLvisible{} points within a vertical plane that contains the target point $T$ 
(Section~\ref{sec:2D_pvp_problem}). Then, we describe some geometrical properties (Section~\ref{sec:2D_pvp_properties}) and introduce a novel algorithm (PVA-2D) to solve PVP-2D (Section~\ref{sec:2D_pvp_algorithm}). This algorithm is the core of our work, as it significantly reduces the computation time of the overall path planner \approach{}. In Section~\ref{sec:2D_pvp_correctness}, we prove the correctness and time complexity of PVA-2D; and in Section~\ref{sec:2D_pvp_extension}, we discuss extensions of the algorithm. Finally, based on PVA-2D, we propose an algorithm (PVA-3D) to solve PVP in 3D scenarios (Section~\ref{sec:3D_pvp}). 

\subsection{PVP-2D Statement}
\label{sec:2D_pvp_problem}

\begin{figure}[t]
    \centering
    \includegraphics[width=0.7\textwidth, page=1]{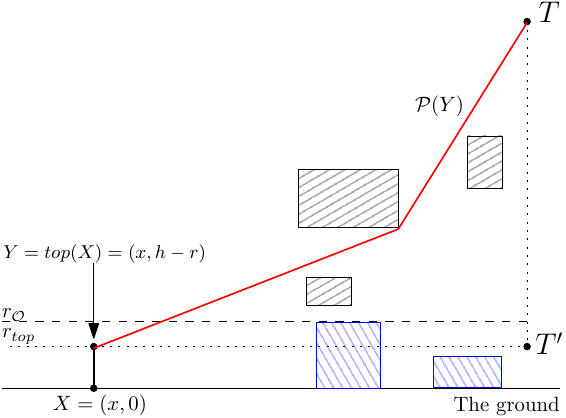}
    \caption{
    An intance of PVP-2D. Aerial obstacles are in gray and ground obstacles in blue. $\mathcal{P}(Y)$ is a CICP from $top(X)$ to $T$. 
    }
    \label{fig:intro}
\end{figure}

After taking off, the motion of the UAV is constrained to the vertical plane $\pi_{X,T}$, hence we address PVP in two dimensions. The intersection of the marsupial system (represented as a cylinder in 3D) with $\pi_{X,T}$ results in a rectangle; and, after the obstacles are enlarged, it becomes a segment of height $h-r$ perpendicular to the ground at position $X$. In addition, the 3D obstacles modeled as cuboids become rectangles. Figure~\ref{fig:intro} shows a representation of the 2D problem.

We first introduce some definitions, followed by the problem statement. Let $v_0, v_1,  \cdots , v_k$ be the vertices, ordered from left to right, of a polygonal chain $\mathcal{P}$ in the plane\footnote{A polygonal chain is defined by an ordered list of its vertices; i.e., $\mathcal{P}=\langle v_0, v_1,  \cdots , v_k\rangle$.}. 

\begin{definition}
 We say that a vertex $v_i$ is a \textit{convex vertex}, $i \neq \{0,k\}$, if $v_i$ is to the right of (or lies on) the directed line from $v_{i-1}$ to $ v_{i+1}$; when all $v_i$ are convex, we say that $\mathcal{P}$ is a \textit{convex polygonal chain}.    
Given a point $p$, we denote by $NE(p)$ the region of the plane to the right and above $p$.  When $v_i \in NE(v_{i-1}), \, i=1, \dots, k,$ we say that $\mathcal{P}$ is an \textit{increasing polygonal chain}.
Finally, a polygonal chain that does not cross any obstacle in the plane is \textit{collision-free}.    
\end{definition}

We denote by $\mathcal{P}(Y)$ a \textit{\textbf{C}ollision-free \textbf{I}ncreasing \textbf{C}onvex \textbf{P}olygonal} chain (CICP) from point $Y=top(X)$ to target $T$, 
and $\mathcal{P}^*(Y)$ represents the minimum-length CICP from $Y$ to $T$.

We propose the following two-dimensional problem, refer to Figure \ref{fig:intro}:

\vspace{0.25cm}

\textbf{PVP-2D:} \emph{Consider three parallel lines in a vertical plane:  $ y=0$,  the ground; $\takeoffline{} \equiv y=h-r$, the take-off line; and $r_\mathcal{O} \equiv  y=h$; and the point $T$ above them. Let $\mathcal{O}$ be the set of $n$ obstacles modeled as isothetic rectangles above the ground. We want to compute the locus of points $Y=top(X)$ in $\takeoffline{}$, with $X=(x,0)$ on the ground to the left of $T$, fulfilling the following conditions:}  

  \begin{enumerate}
      \item[I)] \emph{the marsupial system within the vertical plane located at $X$, —specifically, the vertical segment extending from the lowest endpoint $X$ to a height of $h-r$,
      does not intersect any obstacle in $\mathcal{O}$;}
     \item[II)]  
     \emph{the point $Y$ is \pLvisible{}, which means that there exists a path $\mathcal{P}(Y)$ 
     connecting $Y$ to $T$ with a length not exceeding $L$.
     }
   \end{enumerate}

\vspace{0.25cm}

In order to solve PVP-2D, we divide the set of obstacles $\mathcal{O}$ into two subsets: $\mathcal{O}_a$ and $\mathcal{O}_g$, representing aerial and ground obstacles, respectively. For simplicity, from now on we assume that: 1) the obstacles in $\mathcal{O}_g$ are below the line $r_\mathcal{O}$, while the base of the obstacles in $\mathcal{O}_a$ is placed above the line $r_\mathcal{O}$
; 2) the obstacles are completely contained in the semiplane to the left of the segment $\overline{TT'}$, where $T'$ is the orthogonal projection of $T$ onto \takeoffline{}. 

Let $[(a_i,0), (b_i,0)]$ be the interval of points on the ground covered by the $i$-th ground obstacle.  It is easy to see that all points $X=(x,0)$, such that $x\notin [a_i, b_i]$ for every $i\in [1, 2, \ldots, |\mathcal{O}_g|]$, meet Condition I. From Condition II, 
we need to identify the intervals of \nonpLvisible{} take-off points in $\takeoffline{}$ that are located to the left of $T$. This leads us to the following definition.

\begin{definition}
Let $(A,B)$ be an open interval in \takeoffline, such that $A$ and $B$ are \pLvisible{} points. If each point within $(A,B)$ is \nonpLvisible, then we call $(A,B)$ a maximal \nonpLvisible{} interval, where $A$ and $B$ are named the left and right end-points, respectively.
\end{definition} 

In the following, we exploit several geometric properties that will aid in proposing an algorithm to solve the PVP-2D problem. Specifically, we calculate the set of maximal \nonpLvisible{} intervals contained in  $[Q, T']$, where $Q$ is the point in \takeoffline{} at distance $L$ from $T$ to the left of $T'$.

\subsection{PVP-2D Properties}
\label{sec:2D_pvp_properties}

\begin{figure*}[ht]
    \centering   
    \includegraphics[width=0.9\textwidth]{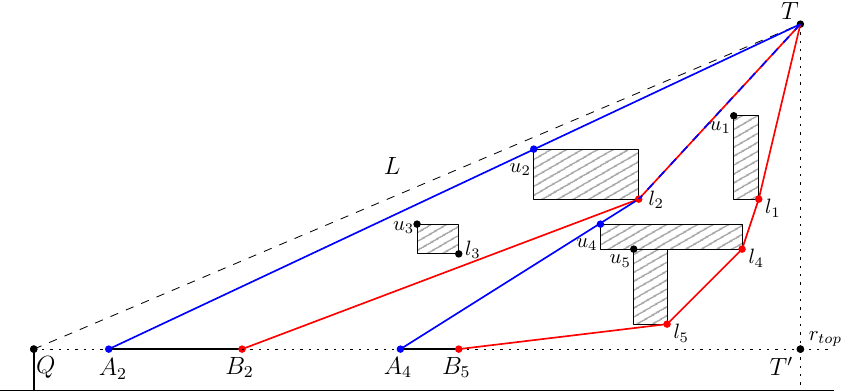}
    \caption{ 
The red and blue points in \takeoffline{} indicate the vertices of the maximal \nonpLvisible{} intervals.
Notice that the red point $l_2$ belongs both to the paths $\mathcal{P}^*(B_2)$ and $\mathcal{P}^*(A_4)$. 
The take-off point $Q$ is the leftmost point that can reach $T$ with a tether of maximum length $L$.
}
    \label{fig:rectangles_solved}
\end{figure*}

For the $i$-th isothetic rectangle in $\mathcal{O}_a$, we denote $u_i$ (resp. $l_i$) its  upper-left (resp. lower-right) vertex (see Figure~\ref{fig:rectangles_solved}). We refer to these points as \textit{critical vertices}. The following properties 
are straightforward to prove:

\begin{lemma}\label{lemma:type1}
Given $A,B \in \takeoffline{}$ and $(A,B)$ a maximal \nonpLvisible{} interval, then there exists $\mathcal{P}^*(A)$, with  $|\mathcal{P}^*(A)|=k > 2$, such that: 1) for all $j\notin \{0, 1, k-1\}$ the vertices $v_j$ of $\mathcal{P}^*(A)$ are the lower-right vertices of some aerial obstacles;  2) $v_1$ is the upper left vertex of an aerial obstacle; 3) $v_0$, $v_1$, and $v_2$ are colinear; and 4) $A$ is the rightmost point so that $\mathcal{P}^*(A)$ contains $v_1$. 
\end{lemma}

\begin{lemma}\label{lemma:type2} 
Given $A,B \in \takeoffline{}$ and $(A,B)$ a maximal \nonpLvisible{} interval, then there exists $\mathcal{P}^*(B)$, such that for each $j\notin \{0,k-1\}$, with $|\mathcal{P}^*(B)|=k > 1$, the vertices $v_j$  of $\mathcal{P}^*(B)$ are the lower-right vertices of some aerial obstacles. In addition, the length of $\mathcal{P}^*(B)$ is $L$.
\end{lemma}

Figure \ref{fig:rectangles_solved} shows the \nonpLvisible{} intervals $(A_2, B_2)$ and $(A_4, B_5)$ for a scenario with five isothetic rectangular obstacles to the left of $T$. The critical vertices $l_i$ and $u_i$ ($i=1, \cdots, 5$) are shown.

\subsection{Polygonal Visibility Algorithm 2D (PVA-2D) }
\label{sec:2D_pvp_algorithm}

Here, we outline the algorithm designed to compute the locus of points as a solution the output to the PVP-2D problem.The algorithm computes a list of \nonpLvisible{} intervals of take-off points in the horizontal line $r_{top}$ within a vertical plane, that is, the complementary set to the one we are looking for. Figure \ref{fig:rectangles_solved} illustrates an example of the locus, which is a set of maximal segments. For a given point $Y$ in the vertical plane, we denote the chain computed by the algorithm as $\mathcal{P}_{alg}(Y)$. The algorithm, referred to as PVA-2D, follows 
the following detailed steps: 

\vspace{0.2cm}

\noindent \underline{{\bf STEP 1:}} \emph{Compute $\mathcal{P}_{alg}(l_j)$ recursively for all lower-right vertices $l_j$.}

\vspace{0.2cm}

Let $l_1, l_2, \cdots, l_n$ be the lower-right vertices of the aerial obstacles arranged in decreasing order of their 
$y$-coordinates; and let $l_0=T$. For each $j \in \{0,1, \cdots, n\}$, we compute, if it exists, a CICP, called $\mathcal{P}_{alg}(l_j)$, using the following two arrays: 

\vspace{0.2cm}

\textbf{(1)} Array $\mathcal{L}$ of chain lengths, where $\mathcal{L}[j]$ stores the length of $\mathcal{P}_{alg}(l_j)$. Initially, $\mathcal{L}[j]=\infty$ for $j\neq 0$, $\mathcal{L}[0]=0$, and the following updating rule is applied:

\begin{equation}\label{eq:B}
		\mathcal{L}[j] = \min_{i\in I_j} \left\{  \mathcal{L}[i] + \|l_i - l_j\|_2 \right\},
	\end{equation}
where $i \in I_j$ iff $\overline{l_jl_i} \cup \mathcal{P}_{alg}(l_i)$ is a CICP. 




\vspace{0.2cm}
	
\textbf{(2)} Array of hooks $\mathcal{D}$, where $\mathcal{D}[j]$ stores $l_i$ with $i\in \{0, \cdots, j-1\}$, whenever $\mathcal{P}_{alg}(l_j)=\langle l_j, l_i, ..., T\rangle$. Initially, $\mathcal{D}[j]=\emptyset$ for all $j$. In Figure \ref{fig:rectangles_solved}, $\mathcal{P}_{alg}(l_5)=\langle l_5, l_4, l_1, T\rangle$.

\vspace{0.2cm}

Notice that updating the value of $\mathcal{L}[j]$ triggers an update in the value of $\mathcal{D}[j]$, so $\mathcal{L}$ and $\mathcal{D}$ are filled simultaneously. The array $\mathcal{D}$ is used to compute $\mathcal{P}_{alg}(l_j)$ during the process.
		
	\medspace

\noindent\underline{{\bf STEP 2:}}  \emph{Compute the  \textit{left end-points} candidates of maximal \nonpLvisible{} intervals.} 

\vspace{0.2cm}

For each upper-left vertex $u_i$, $i\in \{1, \cdots, n\}$, let $l^*_{j_i}$, for some $j\in \{0,1, \cdots, n\}$, be the lower-right vertex such that $\overline{u_il^*_{j_i}} \cup \mathcal{P}_{alg}(l^*_{j_i})$ is a CICP and $\overline{u_il^*_{j_i}}$ has maximum slope (Lemma \ref{lemma:alg-opening} justifies this selection). Let $A^*_{i,j}$ be the intersection point between \takeoffline{} and the line passing through $u_i$ and $l^*_{j_i}$. We create the following two arrays:

\vspace{0.2cm}

\textbf{(3)} Array $\mathcal{A}$ of the \textit{left end-points} candidates:

\begin{equation}\label{eq:A}
		\mathcal{A}[i] = 
		\begin{cases}
   			A^*_{i,j}, & \text{ if } \|A^*_{i,j} - l^*_j\|_2 + \mathcal{L}[j] \leq L \text{ and} \
    \overline{A^*_{i,j}l^*_j} \cap \mathcal{O}_a = \emptyset\\
			\emptyset, & \text{ otherwise. } \\
		\end{cases} 
  \end{equation}

\vspace{0.2cm}
 
\textbf{ (4) } Array of hooks $\mathcal{U}$, where $\mathcal{U}[i]$ stores the point $l^*_j$ if $\mathcal{A}[i]=A^*_{i,j}$. Initially, $\mathcal{U}[i] = \emptyset$ for all values of $i$. 

\vspace{0.2cm}

For simplicity, since there can only be one $A^*_{i,j}$ for each $u_i$, we denote $A^*_{i,j}$ by $A_i$. See the left end-point $A_4$ in Figure \ref{fig:rectangles_solved}. The arrays $\mathcal{L}$ and $\mathcal{D}$ are used to compute the array $\mathcal{A}$.

\medspace

\noindent \underline{{\bf STEP 3:}}  \emph{Compute the  \textit{right end-points} of maximal \nonpLvisible{} intervals.}

\vspace{0.2cm}

\noindent For each $j$, with $\mathcal{L}[j]< \infty$, 
take the point $B_j\in\takeoffline$ to the left of $l_j$ 
such that $\|B_j- l_j\|_2 = L - \mathcal{L}[j]$, if it exists. We create the following array:

\vspace{.2cm}

\textbf{ (5) } Array $\mathcal{B}$ of \textit{right end-points}. Initially, $\mathcal{B}[j]=\emptyset$ for all values of $j$. Then, we apply the update rule $\mathcal{B}[j] = B_j$ if:	
	\begin{enumerate}
		\item[a)]
  $\overline{B_jl_j} \cup \mathcal{P}_{alg}(l_j)$ is a CICP.  
  
\item[b)] There is no $l_k$ so that $\overline{B_jl_k} \cup \mathcal{P}_{alg}(l_k)$ is a CICP with length lower than $L$. Notice that, by definition of $B_j$,  the length of $\overline{B_jl_j} \cup \mathcal{P}_{alg}(l_j)$ is exactly $L$. To illustrate this property, imagine that in Figure \ref{fig:rectangles_solved}, the obstacle labeled as 4 were not present; then the vertex $l_2$ could be used to invalidate $B_5$. 
	\end{enumerate}


\medspace
		
\noindent \underline{{\bf STEP 4:}} \emph{Compute the \pLvisible{} intervals.}

\vspace{0.2cm}

We create the list $\mathcal{H}$ that contains all the points $A_{i}$ and $B_j$ found in the previous steps and sort it from left to right. 
We consider the interval $(A_{i}, B_j)$ as \nonpLvisible{}  when $B_j$ immediately succeeds $A_{i}$ (from left to right) in $\mathcal{H}$. 
If two left end-points $A_{i}$ and $A_{k}$ are consecutive (from left to right), then $A_{i}$ is rejected. 

We create the set $\mathcal{I}$ with the \nonpLvisible{} intervals obtained from $\mathcal{H}$. In the case where $\mathcal{H}$ starts with a right end-point candidate $B_j$, we add to $\mathcal{I}$ the interval $[Q,B_j)$. Finally, the output of PVA-2D, which is the set of \pLvisible{} intervals, is the complement of $\mathcal{I}$ within the interval $[Q,T']$ in \takeoffline.

\subsection{
Correctness and Complexity}
\label{sec:2D_pvp_correctness}
Leveraging the geometric properties of the problem, we can prove the correctness of PVA-2D.
Then, we study its computational cost. 

\begin{remark}\label{remark:alg-pvisible}
    All take-off points $Y\in\mathcal{H}$ (STEP 4 of PVA-2D) are \pLvisible.
\end{remark}

\begin{remark}\label{remark:Larray}
For each $l_j$, $\mathcal{P}_{alg}(l_j) =\mathcal{P}^*(l_j)$ (see Equation~\eqref{eq:B}). Therefore, the array $\mathcal{L}$ stores the length of $\mathcal{P}^*(l_j)$ in the position $j$.    
\end{remark}

From these Remarks, we can prove the following results:

\begin{lemma}\label{lemma:PB}

Let $Y\neq Q$ be a point in \takeoffline. The length of $\mathcal{P}^*(Y)$ is $L$ if and only if $Y\in\mathcal{B}$ (STEP 3 of PVA-2D).
\end{lemma}

\begin{proof}

Let $Y$ be a point in $\takeoffline$, and $\mathcal{P}^*(Y)= \{Y, l_j,\dots,T\}$ be a CICP in which all vertices, except for $Y$ and $T$, are lower-right vertices of aerial obstacles; this chain exists by Lemma \ref{lemma:type2}. 


\sqbox{$\Rightarrow$} Consider $\mathcal{P}^*(Y)$ of length $L$. By Remark \ref{remark:Larray}, there exists $\mathcal{P}^*(l_j)$ such that: 
    $$\mathcal{P}^*(Y)=\{Y,l_j\} \cup \mathcal{P}^*(l_j) = \overline{Yl_j} \cup \mathcal{P}_{alg}(l_j)$$ 
    This implies that $\|Y-l_j\|_2 + \mathcal{L}[j] = L$. Therefore, the point $Y$ is considered by PVA-2D in STEP 3 and stored in $\mathcal{B}[j]$.



\sqbox{$\Leftarrow$} Conversely, suppose $Y = \mathcal{B}[k]$ be selected in STEP 3 of PVA-2D. Assume that the length of $\mathcal{P}^*(Y)$ is less than $L$. In this case, $\mathcal{P}^*(Y)$ would be shorter than the CICP $\{Y, l_k\} \cup \mathcal{P}_{alg}(l_k)$, where $k\neq j$. However, by Remark \ref{remark:Larray}, there exists $\mathcal{P}^*(l_j)$ such that: 
    $$\mathcal{P}^*(Y)=\{Y, l_j\} \cup \mathcal{P}^*(l_j) = \{Y, l_j\} \cup \mathcal{P}_{alg}(l_j) $$
Therefore, the length of $\{Y, l_j\} \cup \mathcal{P}_{alg}(l_j)$ must be less than the length of $\{Y, l_k\} \cup \mathcal{P}_{alg}(l_k)$. Consequently, the point $Y$ would be discarded as the right end-point of a maximal \nonpLvisible{} interval in STEP 3 (b). This contradicts the assumption that the length of $\mathcal{P}^*(Y)$ is less than $L$, and the result follows.
\end{proof}

\begin{lemma}\label{lemma:alg-opening}
 If $(A,B)$ is a  maximal \nonpLvisible{} interval found in list $\mathcal{H}$ at STEP 4 of PVA-2D, then $A$ is the first \pLvisible{} point to the left of $B$ in \takeoffline.
\end{lemma}

\begin{proof}

According to Lemma \ref{lemma:PB}, $\mathcal{P}^*(B)$ and $\mathcal{P}_{alg}(B)$ coincide and both have length $L$. Let $Y$ be the first \pLvisible{} point in \takeoffline{} to the left of $B$. As the length of $\mathcal{P}^*(B)$ is $L$, if $Y$ exists, then $(Y,B)$ is a maximal \nonpLvisible{} interval.
 
\begin{figure}[t]
\centering
\includegraphics[width=0.6\textwidth]{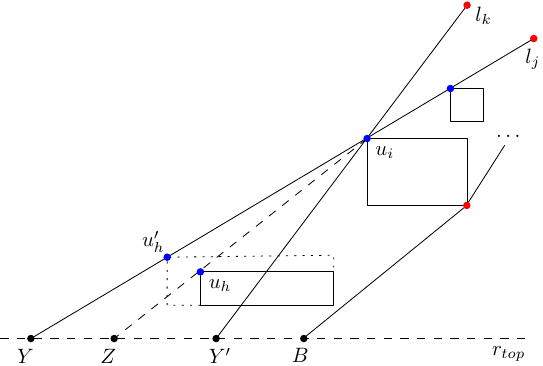}
\caption{Proof of Lemma \ref{lemma:alg-opening}. Segment $\overline{u_il_j}$ is the one with greatest slope if $Y$ is the first \pLvisible{} point to the left of $B$.}
\label{fig:slopes_proof}
\end{figure}

Now, let $u_i$ and $l_j$ the second and third vertices of the CICP $\mathcal{P}^*(Y)$ as stated in Lemma \ref{lemma:type1}. Without loss of generality, we assume that $u_i$ is the lowest upper-left vertex if there are several. Refer to Figure \ref{fig:slopes_proof}. We can prove that, if $Y$ is the first \pLvisible{} to the left of $B$, then the segment $\overline{u_il_j}$ has the highest slope among all pairs $(u_i,l_k)$ so that $\overline{u_il_k}\cup\mathcal{P}_{alg}(l_k)$ is a CICP. 
Suppose, on the contrary, that there exists a segment $\overline{u_il_k}$
where $\overline{u_il_k}\cup\mathcal{P}_{alg}(l_k)$ is a CICP and
$\overline{u_il_k}$
has a greater slope than $\overline{u_il_j}$, as shown in Figure \ref{fig:slopes_proof}; and let $Y'$ be the intersection point between \takeoffline{} and the line containing $\overline{u_il_k}$. As there is no \pLvisible{} point in \takeoffline{} between $Y$ and $B$, there exits an obstacle below $O_i$ that intersects the segment $\overline{Y'u_i}$. Moreover, there exists an obstacle that necessarily touches the segment $\overline{Yu_i}$. In this case, $u_i$ is not the lowest left-upper vertex of $\mathcal{P}^*(Y)$, leading to a contradiction. As PVA-2D selects the pair $\overline{u_il_j}$ with the greatest slope, $Y=A$ and the Lemma follows.
\end{proof}

\begin{theorem}\label{theo:correctness}
Let $Y$ be a point in the interval $[Q,T']$ in \takeoffline. The point $Y$ is \nonpLvisible{} if and only if $Y$ belongs to an interval in the set $\mathcal{I}$ found in STEP 4 of PVA-2D.
\end{theorem}

\begin{proof}

\sqbox{$\Rightarrow$} Let $Y$ be a point in \takeoffline{}. If $Y$ belongs to an interval $(A,B)$ of $\mathcal{I}$, then according to Lemma \ref{lemma:alg-opening}, $A$ is the first \pLvisible{} point to the left of $B$ in \takeoffline. If $Y$ belongs to the interval $[Q,B)$ of $\mathcal{I}$, then as a consequence of Lemma \ref{lemma:alg-opening}, there are no \pLvisible{} points to the left of $B$ in \takeoffline. In either case, $Y$ is a \nonpLvisible{} point.

\sqbox{$\Leftarrow$} For the reverse implication: let $Y$ be a \nonpLvisible{} point. Let $B$ be the first \pLvisible{} point in \takeoffline{} to the right of $Y$. Notice that $\mathcal{P}^*(B)$ must have length $L$; therefore, according to Lemma \ref{lemma:PB}, PVA-2D finds $B$ as the right end-point of a maximal \nonpLvisible{} interval in STEP 3. Now, let $A$ be the first \pLvisible{} point in \takeoffline{} to the left of $Y$. If $A$ exists, then according to Lemma \ref{lemma:alg-opening}, the algorithm selects $(A,B)$ as a maximal \nonpLvisible{} in STEP 4. If $A$ does not exist, then $B$ would be the first point in the list $\mathcal{H}$, and the interval $[Q,B)$ would be added to $\mathcal{I}$ in STEP 4. In either case, $Y$ belongs to an interval in $\mathcal{I}$, and the result follows.
\end{proof}

Now, we establish the computational complexity of PVA-2D, assuming that a visibility graph is used for determining the collision-free paths. Notice that computing the visibility graph of a collection of disjoint polygons in the plane with a total of $n$ edges takes $O(n^2)$ time \citep{asano1985visibility}. 

\begin{theorem}\label{theo:complexity}
PVA-2D takes $O(n^2)$ time to compute the set of \pLvisible{} intervals.
\end{theorem}

\begin{proof}

As a pre-processing step, we build a visibility graph between $T$ and the aerial obstacles in the vertical plane. The visibility graph allows us to check in $O(1)$ time whether all two critical vertices are visible. As we consider $n$ rectangular obstacles and the point $T$, the total number of vertices in the graph will be $4n+1$, and this process takes $O(n^2)$ time. 

(STEP 1) We sort the lower-right critical vertices $l_i$ in decreasing order of height in $O(n\log n)$ time. Then, the arrays $\mathcal{L}$ and $\mathcal{D}$ of PVA-2D can be calculated with two nested loops on the sorted $l_i$ vertices, taking $O(n^2)$ time. 

(STEP 2) The candidates for left end-point of maximal \nonpLvisible{} interval can be calculated in $O(n^2)$ time as follows. 
The key observation is that for each upper-left vertex $u_i$, we can compute its associated left endpoint in $O(n)$ time. First, compute the $l_j$ vertex so that $\{u_i\}\cup\mathcal{P}_{alg}(l_j)$ is a CICP and $\overline{u_il_j}$ has the maximum slope. Let $A_i$ be the associated point in \takeoffline{} such that $u_i\in \overline{A_il_j}$. 
Second, if $\overline{A_il_j}$ does not collide with any obstacle, then $A_i$ is a candidate to left end-point of interval and we set $\mathcal{U}[i]=A_i$; else, according to Lemma \ref{lemma:alg-opening}, we do not need to check for more pairs $(u_i,l_k)$ and we can avoid quadratic time at this step. Therefore, as we require only $O(n)$ time for each $u_i$, the total time to fill the array $\mathcal{A}$ is $O(n^2)$ time. This aligns with the overall complexity of constructing the visibility graph, ensuring that the algorithm remains efficient.

(STEP 3) Afterward, to fill the array $\mathcal{B}$, we must search for all points $B_j$ in \takeoffline{} such that $\{B_j\} \cup \mathcal{P}_{alg}(l_j)$ forms a CICP of length $L$. For each $l_j$, we calculate the point $B_j$ in $O(1)$ time. We then add all candidates $B_j$ to the precomputed visibility graph, taking $O(n^2)$ in the worst case. The updated visibility graph enables us to check whether there exists a lower-right vertex $l_k$  such that $\{B_j\} \cup \mathcal{P}_{alg}(l_k)$ forms a CICP of length lower than $L$. If such a $l_k$ exists, then $B_j$ is discarded as a candidate. Since there are at most $n$ candidates $B_j$, the $\mathcal{B}$ array can be filled in $O(n^2)$ time. 

(STEP 4)  Finally, the list $\mathcal{H}$ is created in $O(n)$ time by joining the points $A_i$ and $B_j$. We sort this list in $O(n\log n)$ time and the maximal \nonpLvisible{} intervals in the $\mathcal{I}$ array can be computed with a left-to-right sweep over $\mathcal{H}$ in $O(n)$ time. The output of the algorithm is the complement of $\mathcal{I}$ in the interval $[Q,T']$.
\end{proof}

Notice that, since the number of vertices per obstacle is constant, the complexity of the algorithm depends entirely on the number of obstacles. Another observation is that if we are given the sorted list of \pLvisible{} intervals, determining whether
a query point $Y$ in \takeoffline{} is \pLvisible{} can be answered in $O(\log{n})$ time using binary search.

\subsection{Extensions}
\label{sec:2D_pvp_extension}

This Section discusses how our algorithm could be extended by relaxing some of the initial assumptions:

\begin{itemize}

    \item \textbf{Ground obstacles above the $r_\mathcal{O}$ line.}  
Let $O$ be an obstacle 
on the ground with height greater than $h$. As with the rest of the ground obstacles, we first use $O$ to discard the intervals of points on the ground where the UGV would collide. After that, $O$ should be considered as another aerial obstacle for the rest of the algorithm; but notice that there is no need to consider its lower-right vertex as part of any CICP in the PVA-2D, as that vertex is on the ground. 

    \item \textbf{Aerial obstacles below $r_{\mathcal{O}}$.} 
    
Let $O$ be an aerial obstacle below $r_{\mathcal{O}}$. In this case, $O$ can also be considered as a ground obstacle, since 
marsupial system 
cannot pass below it. Thus, we have to use this obstacle in the preprocessing step to discard the intervals where the UGV cannot enter. Besides, this obstacle should also be considered as another aerial obstacle in the rest of the algorithm. Note that a special case could occur: a polygonal line from a point $Y\in\takeoffline{}$ to $T$ could be tangent to the base of $O$ and touch both its lower (left and right) vertices that support the bottom face. Consequently, the array $\mathcal{L}$ used in PVA-2D to store the length of $\mathcal{P}^*$ from each lower-right vertex to $T$, can also be used to store the length of $\mathcal{P}^*$ from each lower-left vertex to $T$ without altering the time complexity of $O(n^2)$ of STEP 2. Furthermore, since now there could be polygonal lines below $r_\mathcal{O}$, ground obstacles must also be considered in tether collision check.

\item \textbf{Obstacles on both sides of $T$. } Let us call central obstacles those intersecting the segment $\overline{TT'}$, where $T'$ is the orthogonal projection of $T$ on the ground. The problem can be solved independently in each halfplane defined by $\overline{TT'}$. The only two updates we need to make to include these obstacles to PVA-2D are: (1) do not calculate the CICP $\mathcal{P}_{alg}(l_c)$ in STEP 1 for all lower-right vertices of central obstacles $l_c$ and, (2) consider the projection point $T'$ as the right-endpoint of a maximal \nonpLvisible{} interval in STEP 4.


    \item \textbf{Different types of obstacles.}  
The algorithm can be easily extended to non-rectangular obstacles. In that case, CICPs of minimum length could be tangent to many sides of the obstacles. Basically, we have a convex chain that plays the role of the vertex $u_i$. Therefore, in the array $\mathcal{L}$, we would need to store the length of $\mathcal{P}^*(v)$, for each vertex $v$ of each obstacle. 
    
\end{itemize}

The complexity of PVA-2D depends entirely on the total number of vertices. It is easy to see that the above extensions can be solved in the same $O(n^2)$ time, with $n$ being the total number of vertices of the obstacles.


\subsection{ Polygonal Visibility Algorithm 3D (PVA-3D)}
\label{sec:3D_pvp}

The PVA-2D algorithm can be used to 
design an approximation algorithm, PVA-3D, for the general polygonal visibility problem in 3D, where the goal is to compute the locus of all \pLvisible{} points in the plane $\pi_{top}$.
The idea is to consider a beam of $p$ planes, passing through $T$ and perpendicular to the ground, and use PVA-2D in each plane. Clearly, the more planes considered, the better approximation of the \pLvisible{} region becomes.

\begin{figure}
\centering
\includegraphics[width=0.6\textwidth]{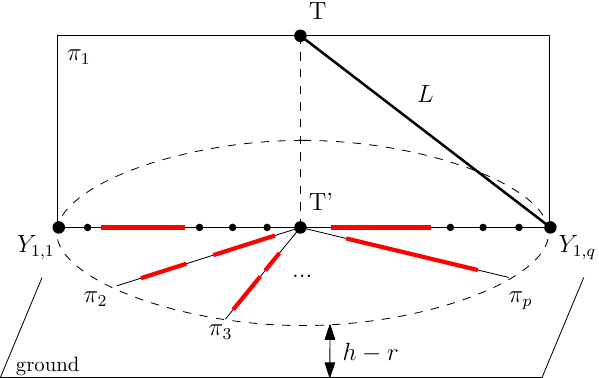}
\caption{Approximated solution for PVP-3D. For each plane $\pi_i$, a discrete set $\mathcal{Y}_i$ is considered as feasible take-off points. The red segments indicate \nonpLvisible{} intervals in each vertical plane.} 
\label{fig:c-vis}
\end{figure}

Consider the circle on $\pi_{top}$ centered at $T'$ with radius $r=\sqrt{L^2-\|T-T'\|_2^2}$, which contains all feasible take-off points. We uniformly divide the space into a set $\mathcal{\pi}=\{\pi_1, \pi_2, \dots, \pi_p\}$ of $p$ vertical planes containing $T$ and the output of applying PVA-2D in these planes gives a discrete approximation to the 2D region of \pLvisible{} points.Figure~\ref{fig:c-vis} illustrates the aproximation.

\section{MArsupial Sequential Path-planning Approach (\approach)}\label{sec:opt}

In this Section, we present our strategy for solving the SMPP (Shortest Marsupial Path Problem) in a 3D scenario. Our approach, \approach{}, involves navigating the UGV from its starting position $S$ to a designated point $X$. Once the UGV reaches this location, the UAV is deployed to successfully reach the aerial target $T$.

Assume that a solution of PVA-3D is provided as described in Section \ref{sec:3D_pvp}. In each plane, we consider a discrete set $\mathcal{Y}_i=\{Y_{i,1},\dots, Y_{i,q}\}$ of $q$ take-off candidate points uniformly distributed along the \pLvisible{} intervals, where $Y_{i,j}$ is the $j^{th}$ candidate in the $i^{th}$ vertical plane, for $i=1,\dots, p$. See Figure~\ref{fig:c-vis} for an illustration of the process. Let $\mathcal{P}^*(Y_{i,j})$ be the minimum length CICP path for the UAV between $Y_{i,j}$ and $T$.
Recall that, for each plane $\pi_i$, PVA-2D stores in the array $\mathcal{L}$ the length of $\mathcal{P}_{alg}(l_k)$, for each lower-right vertex $l_k$. 
With this information, we can compute $\mathcal{P}^*(Y_{i,j})$ for the point $Y_{i,j}$ in $O(n^2)$ time by checking each possible segment $\overline{Y_{i,j}l_k}$ for collisions and keeping the one that defines the shortest path. Therefore, we compute the discrete set of \pLvisible{} take-off points in $O(pqn^2)$ time, where $pq$ represents the maximum number of candidate points.


Let $G=(V,E)$ be the visibility graph on the ground
where the set of vertices $V$ is defined by the starting point of the UGV, $S$, the vertices of the projection on the ground of the obstacles in $\mathcal{O}_g$, and the set of the UGV positions $X_{i,j}$ corresponding to the \pLvisible{} $Y_{i,j}$ points, that is, $Y_{i,j}=top(X_{i,j})$. 

An edge determined by a
pair of vertices $v_i,v_j$ in $V$ is included in $E$ if the UGV can travel from $v_i$ to $v_j$ in a straight line without collision (recall that the UGV is modeled by a vertical segment according to the model explained in Section \ref{sec:model}). %
As obstacles are cuboids, their projection on the ground has 4 vertices; therefore, the number of elements in the set $V$ is at most $4n + pq + 1$. 
Note that $G$ could be partially computed in a preprocessing step when the obstacles in the space are known in advance, which would ease the overall computational effort for problem instances involving multiple target points. 

The optimization problem SMPP aims to minimize the total path length for both the UGV and the UAV. To achieve this, we introduce an additional vertex $T'$
into the graph $G$. This vertex, $T'$, is connected to each candidate point $X_{i,j}$, where the edge weight between $X_{i,j}$ and $T'$ corresponds to the optimal path length of the UAV, $\mathcal{P}^*(Y_{i,j})$. This ensures that the optimization accounts for both the UGV's ground path and the UAV's aerial path, ultimately minimizing the combined travel distance.
Note that any path from $S$ to $T'$ in the augmented graph contains the sum of the path lengths of both vehicles. 
With the visibility graph at hand, we can apply a graph search approach (such as, for example, Dijkstra's algorithm) to find the shortest path from $S$ to $T'$. 

The computational complexity of \approach{} is related to the size of the graph $G$. As $V$ contains at most $4n+pq+2$ vertices (considering $S$ and $T'$), Dijkstra's algorithm spends $O((n+pq)^2)$ time. As a result, the total time complexity of \approach{} is 
$O(pq(n^2+pq))$ time, where $O(pqn^2)$ is the cost of computing the \pLvisible{} take-off points. Notice that if $n$ is sufficiently large and the parameters $q$ and $p$ are considered constants, the overall time complexity of the MASPA strategy becomes quadratic. 

\section{Controllable Loose Tethers}\label{sec:extension}

We now introduce a variant of the MASPA planner designed to efficiently solve the SMPP in scenarios involving a loose tether.
In this case, we assume that the marsupial system carries a device to control the length of the tether (with a maximum length of $L$). We introduce a new concept: a take-off point $Y=top(X)$ is \emph{catenary-visible}, \cLvisible{} for short, if there exists a collision-free loose tether between $Y$ and $T$ that is modeled by a catenary curve with length equal to or lower than $L$. Then, we redefine the PVP problem for catenary curves.

\vspace{0.25cm}
\textbf{Catenary Visibility Problem (CVP):} \emph{Compute
the locus of the \cLvisible{} points in $\pi_{top}$}.

\vspace{0.25cm}
Notice that, when we consider catenaries, MLTP (Minimum Length Tether Problem) retains its original formulation. However, the procedure to obtain the minimum length tether requires modification to account for feasible catenary curves. 
To efficiently solve CVP and MLTP for loose tethers, we use the following result in 2D:

\begin{lemma}\label{lemma:c-vis}
    If a point $Y \in r_{top}$ is \nonpLvisible, then it is also not \cLvisible{}.
\end{lemma}

\begin{proof}
    Let $Y$ be a \cLvisible{} point; then there exists a catenary of length equal to or less than $L$ connecting $Y$ and $T$. That controllable tether could be tensioned until it transforms into a polygonal line of length equal to or less than $L$, and the result follows.
\end{proof}
As we did in Section~\ref{sec:3D_pvp}, we can generate a discrete set of $p$ vertical planes passing through $T$ and a discrete set of $q$ take-off candidate points in each plane along the \pLvisible{}-intervals. Notice that, according to Lemma~\ref{lemma:c-vis}, the \nonpLvisible{} points cannot be \cLvisible{} either. 
For each of the candidate take-off points, $Y_{i,j}$, we use an additional optimization algorithm to find the collision-free catenary with minimum length from $Y_{i,j}$ to $T$. 

The computation of the collision-free catenaries of minimum length from a take-off point to $T$ is out of the scope of this work; see, for instance, \citep{lopez2006computation} and \citep{aristizabal2023solution} for some related papers.

However, in order to conduct experiments to test the \approach{} strategy, we approximate the MLTP by selecting a discrete number of tether lengths to check, with values between $\overline{Y_{i,j}T}$ and $L$, which are the minimum and maximum possible lengths, respectively. Given two anchor points and the length, we can calculate the equation of the catenary curve and use it to compute the collisions between a catenary curve and the $n$ rectangular obstacles can be checked in $O(n)$. Assuming that the computation of a single catenary can also be performed in $O(1)$ time\footnote{We assume a computational model in which a catenary, a plane transcendental curve defined by two anchor points and a given length, can be computed in $O(1)$ time.}, and considering a number of $c$ catenary lengths to test from a point $Y_{i,j}$, the computational time required to find the tether with minimum length connecting $Y_{i,j}$ and $T$ is $O(cn)$ time. Therefore, the total time to calculate the catenary with minimum length from a set of $q$ feasible take-off points in a vertical plane becomes $O(n^2 +cqn)$, where $O(n^2)$ is the cost of solving the PVP-2D algorithm in the plane. After doing this process for every vertical plane, the total cost is $O(pn(n + cq))$ time.    

Having the length of the minimum tether for the considered take-off points, the \approach{} strategy can be used as we did in the case of taut tethers, i.e., by applying Dijkstra's algorithm over an augmented visibility graph of obstacles on the ground. Therefore, the total time complexity required will be $O((n+pq)^2+pn(n + cq))$ time, where $p$ is the number of vertical planes, $q$ is the number of take-off candidate points in each plane, and $n$ is the total number of vertices of the obstacles. Since the parameters $p$, $q$, and $c$ do not depend on the input, the computation time is bounded by $O(n^2)$ time.
A study on how to tune the parameters $p$ and $q$ is provided in Section \ref{sec:parameter_study}.

\section{Parameters Setting} \label{sec:parameter_study}

A series of computational experiments to analyze the effects of the key parameters in \approach{} 
is described in this Section. All experiments were carried out using a version of \approach{} coded in Python 3.10.7 on a CPU with a $3.20~GHz$ processor and $16~GB$ RAM.

We study the parameters $p$, $q$ which represent the number of planes and the number of take-off points per plane to be considered in PVA-2D, respectively; additionally, we test the parameter $c$, which represents the number of catenary lengths to be used in the approximation of the problem MLTP. To achieve this, we constructed a set of 250 randomly generated scenarios to test and run the \approach{} algorithm.
Each scenario is contained in a box of shape $50\times 50\times 40~m^3$. Within the box, we randomly generate $10$ disjoint obstacles on the ground and $15$ disjoint obstacles in the air, where each obstacle is a cuboid
with side lengths of 5 meters, resulting in a volume of 125 cubic meters. We located the starting point $S$ at a corner of the box and randomly generated the target point $T$ of the UAV in the air, ensuring that the minimum height of $T$ is $25~m$. We also ensured that the points $S$ and $T$ where placed in obstacle-free positions. Finally, we consider the height of the marsupial system to be $h=1.5~m$ and the UAV's radius to be $r=0.5~m$.

In our experiments, we considered a tether of $50~m$ of maximum length. Since that the minimum height of $T$ is $25~m$, the length of the shortest possible catenary connecting a take-off point and $T$ is $25-(h-r)=24~m$. Therefore, we can estimate the maximum error related to the selection of the parameter $c$ in advance. Recall that, in the MLTP, we select $c$ catenary lengths uniformly distributed between the minimum and maximum possible lengths, that is, $24~m$ and $50~m$, respectively, in our experiments. 
Consequently, if we do not consider any obstacles, given the shortest catenary from a take-off point to $T$, the closest catenary considered by the algorithm has a maximum difference in length of: 
$$\frac{50-24}{2c}~m=\frac{26}{2c}~m$$ 

We then use a fixed value of $c=26$ in all experiments, ensuring that the maximum error of length 
is $0.5~m$ when collision is disregarded. The problem of finding a  approximation with guarantee or an optimal solution for MLTP with catenaries is beyond the scope of this work.

With the parameter $c$ fixed, we execute \approach{} in each scenario taking all combinations of the parameter $p$ in the set $\{4,8,16,32\}$ and the parameter $q$ in the set $\{10,20,30,40\}$. Ignoring the presence of obstacles, for the maximum values $p=32$ and $q=40$, the distance between any point in $\pi_{top}$ and its closest candidate take-off point computed using $p$ and $q$ is less than $2.5~m$.

In the experiments, we assess the quality of the results provided by \approach{} using two metrics:

\begin{itemize}

\item \textbf{Total Length (TL)}: The total mission length is the combined sum of the UGV's path and the UAV's flight path.

\item \textbf{Execution Time (ET)}: The total time taken to run \approach.

\end{itemize}

Table \ref{tab:metrics} presents the results of \approach{} in the random scenarios for the TL and ET metrics.
For each metric, we report the mean and standard deviation calculated in all scenarios.
 
\begin{table}[H]
\caption{The mean and standard deviation values for the TL metric (in meters) and the ET metric (in seconds) across the set of random scenarios.}
\begin{center}
\begin{tabular}{ | c || c | c | c | c | } 

  \hline
  \backslashbox[8mm]{$p$}{$q$}   & 10  & 20  & 30  & 40   \\ 
  \hline
  \hline
  4 
  & $65.5\pm6.6$ & $64.9\pm6.4$ & $64.9\pm6.5$ & $67.0\pm6.6$ \\
  \hline
  8 
  & $63.4\pm7.1$ & $63.0\pm6.8$ & $63.1\pm6.9$ & $62.9\pm6.8$ \\
  \hline
  16 
  & $62.3\pm7.0$ & $62.1\pm7.0$ & $\mathbf{62.0\pm7.0}$ & $62.1\pm6.9$ \\
  \hline
  32 
  & $61.7\pm6.9$ & $61.6\pm6.9$ & $61.7\pm7.0$ & $61.6\pm7.0$
  \\
  \hline
\end{tabular}\label{tab:metrics}

\vspace{.5cm}

\begin{tabular}{ | c || c | c | c | c |  } 

  \hline
  \backslashbox[8mm]{$p$}{$q$} & 10  & 20  & 30  & 40  \\ 
  \hline
  \hline  
  4 
  & $0.3\pm0.1$ & $0.4\pm0.1$ & $0.5\pm0.1$ & $0.6\pm0.1$ \\
  \hline
  8 
  & $0.6\pm0.1$ & $0.8\pm0.2$ & $1.0\pm0.2$ & $1.3\pm0.3$ \\
  \hline
  16
  & $1.2\pm0.2$ & $1.6\pm0.4$ & $\mathbf{2.1\pm0.5}$ & $2.6\pm0.6$
 \\
  \hline
  32
  & $2.4\pm0.5$ & $3.3\pm0.7$ & $4.2\pm0.9$ & $5.2\pm1.1$
 \\
  \hline
\end{tabular}
\end{center}
\end{table}

\begin{figure}[t]
\centering

\includegraphics[width=0.6\textwidth]{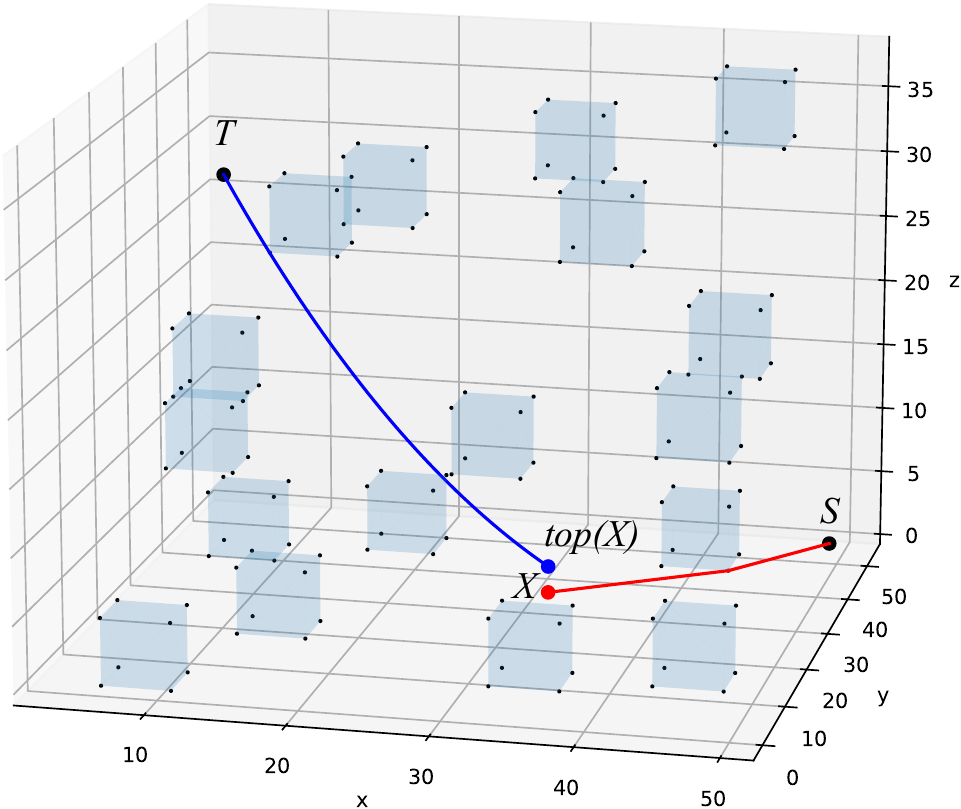}
\caption{Output of the planner \approach{} in a random scenario with $p=16$ and $q=30$. The marsupial system follows the ground path (red) from the starting point $S$ to the point $X$ avoiding obstacles. Then, from $top(X)$, the UAV is deployed and follows the collision-free aerial path (blue) with the shape of a catenary towards the target point $T$. In \approach{}, the point $X$ is selected to minimize the sum of the lengths of the ground and aerial paths.}
\label{fig:exp1}
\end{figure}

The design of efficient algorithms for the planning of a marsupial system is a challenge today, as observed by Mart\'inez-Rozas et al.~\citep{martinez-rozas_icra21}. The authors report that computing optimal marsupial paths can take up to 40 seconds. From our results in Table~\ref{tab:metrics}, we can confirm the intuitive notion that higher values of the parameters $p$ and $q$ result in shorter path lengths but longer execution times. In particular, for parameter values $p=16$ and $q=30$, the mean TL metric differs by less than $0.5~m$ from the best results achieved with the highest parameter values. At the same time, this parameter selection offers an acceptable performance in terms of the ET metric, with a mean of less than $3~s$. Figure \ref{fig:exp1} illustrates the resulting shortest UGV and UAV paths generated by \approach{} in a random scenario using $p=16$ and $q=30$.   

\section{Experimental Evaluation} \label{sec:experiments}

In this Section, we evaluate the application of \approach{} in realistic scenarios and compare its performance with that of a baseline method~\footnote{As far as we know, no planner currently addresses sequential path planning with the UGV first and then the UAV.}. The path planners are compared in two realistic scenarios where obstacles are selected in a structured way. 
Furthermore, we measure the effect of the visibility module (PVA) in \approach{}.

\subsection{RRT* Baseline} 

Due to the lack of studies on sequential marsupial planning, we implement a baseline method based on the well-known RRT* algorithm for path planning~\citep{karaman2011sampling,noreen2016optimal} and compare the results with \approach. The RRT* algorithm is a provably asymptotically optimal method that randomly samples feasible states in the space and connects them into a tree graph, such that the edges of the graph minimize a certain cost function. RRT* is popularly used in problems with dynamic constraints to create the shortest path between initial and ending states while avoiding collisions with obstacles.

For the sake of comparison,  we utilize RRT* to generate collision-free paths on the ground from the starting point $S$ for the UGV. In this scenario, each node in the RRT* tree represents a position on the ground that the UGV can reach without collisions. For each node $X$ on the ground, we check if the point $top(X)$ is \cLvisible{} and, if so, compute the minimum aerial path to $T$ using the CVP algorithm described in Section \ref{sec:extension}. Finally, we select the node $X$ in the tree that minimizes the sum of the path length from $X$ to $S$ and the path length from $top(X)$ to $T$. If no such point exists, we conclude that there is no feasible solution to the problem.

As a stopping parameter in the RRT* algorithm, we consider the maximum resolution time. This parameter defines how long the RRT* algorithm runs to create random points. The longer the algorithm runs, the more random points are generated, increasing the likelihood of obtaining a good solution. However, a poor selection of this parameter can lead to suboptimal solutions or, in some cases, to no solution at all.

\subsection{Comparison} 

\begin{figure}
\centering
\includegraphics[width=0.6\textwidth]{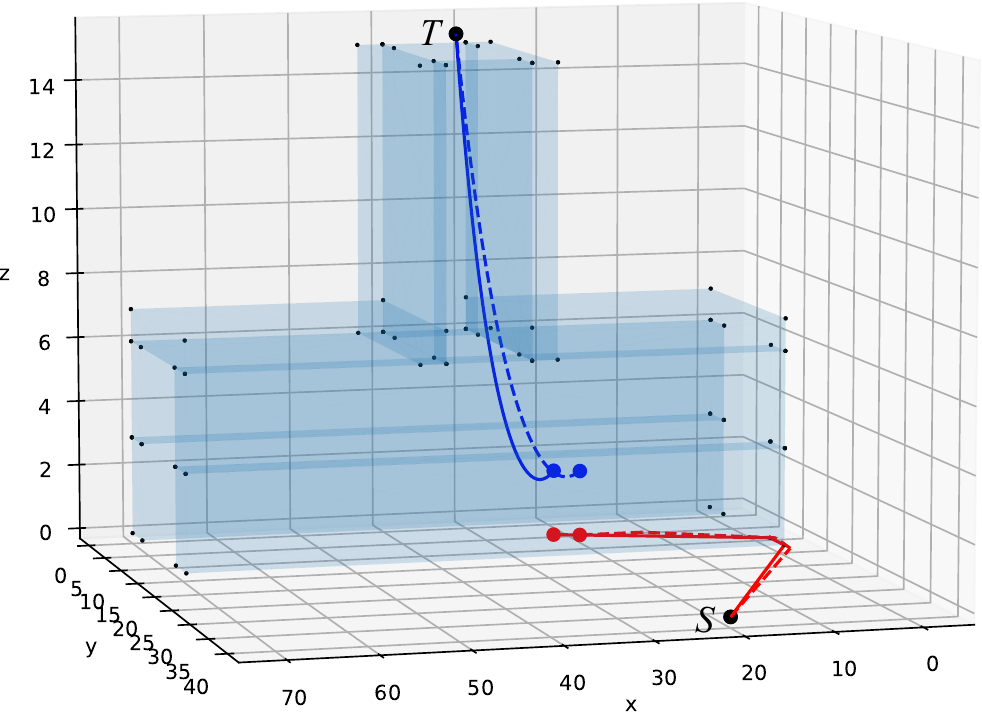}
\caption{Resulting path in scenario S1 (Fireplace) for the \approach{} strategy with $p=16$ and $q=30$, compared to the RRT* baseline executed for 20 seconds. The dashed lines represent the ground and aerial paths retrieved by the RRT* algorithm.}
\label{fig:tunnel}
\end{figure}

\begin{figure}
\centering
\subfloat[\centering]{{\includegraphics[width=0.65\textwidth]{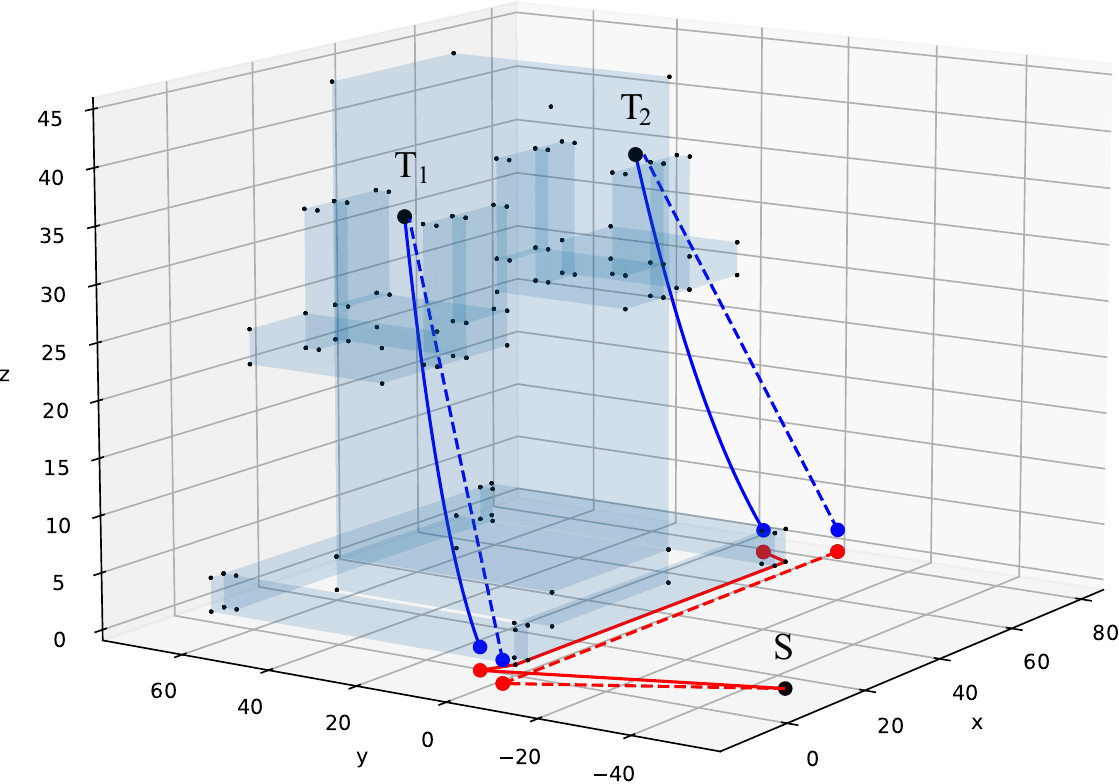} }}%
\qquad
\subfloat[\centering]{{\includegraphics[width=0.5\textwidth]{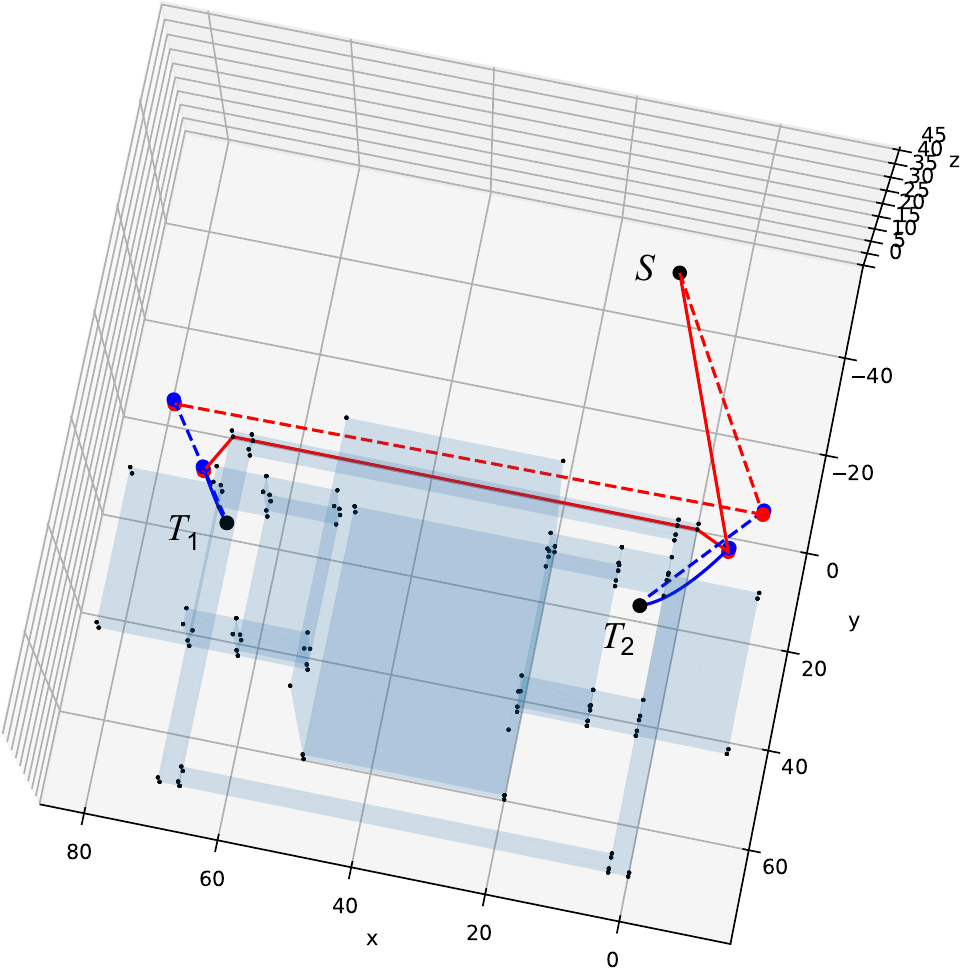} }}%
\caption{(a) Resulting paths in scenario S2 (Balconies) for the \approach{} strategy with $p=16$ and $q=30$, compared to the RRT* baseline, executed for 20 seconds. The dashed lines represent both the ground and aerial paths retrieved by the RRT* algorithm. (b)Top view.}
\label{fig:building}
\end{figure}

We compare the baseline approach based on RRT* with \approach{} in the following two realistic scenarios:  

\begin{itemize}
    \item S1 (Fireplace): This scenario involves a system of walls that simulate a fireplace. The target $T$ is located above the fireplace hole. The marsupial system must enter the enclosure and deploy the UAV through the fireplace to reach $T$, Figure \ref{fig:tunnel}. 
    
    \item S2 (Balconies): This scenario involves a building with two balconies, each occupied by a person who needs help. The marsupial system must plan a path to reach both balconies sequentially. Additionally, there is a designated forbidden area around the building that the UGV must avoid, as illustrated in Figure \ref{fig:building}. This scenario is challenging because the UAV can access the target points only through a small gap in the balconies, as shown in Figure \ref{fig:building}~(c).
\end{itemize}

In both scenarios, we use the \approach{} strategy with parameter values $p=16$ and $q=30$, selected based on the parameter study in Section~\ref{sec:parameter_study}. For RRT*, we set the maximum resolution time parameter to 20 seconds. This duration was chosen because lower values of this parameter resulted in poor solutions and, in some cases, no solutions were found.

In scenario S2, the goal is to compute a path for the marsupial robotics system to sequentially visit two targets, and we use the planners twice, consecutively and independently. First, we compute the optimal path from the starting point to the first target. After reaching it, we assume that the UAV returns to the UGV by retracing its collision-free aerial path. Finally, we plan a path from the UGV's current position to the second target. This scenario is particularly challenging because finding optimal solutions requires the UAV to navigate through the gaps beneath each balcony. In contrast, a suboptimal approach tends to take a longer route that flies directly to the target, passing above the balconies.

Finally, in order to validate the use of the visibility module, we include a third approach, which is the \approach{} strategy without the visibility algorithm, referred to as \approach$^-$. This variant does not use PVA to filter \noncLvisible{} points. Instead, it considers candidates for take-off points that are uniformly located along the straight line $r_{top}$. Table \ref{tab:lengthS1S2} presents the results of the comparison between the three planners in terms of total path lengths. Both the \approach{} and \approach$^-$ strategies are deterministic, which means that they will produce the same output given the same input and parameters. In contrast, the RRT* baseline incorporates a random component, so we run the algorithm 10 times and report the mean and standard deviation of the results in the table. As shown, \approach{} outperforms the RRT* baseline after 20 seconds of execution.

\begin{table}[H]
\caption{Mean and standard deviation values (in meters) for the TL metric across the different planners.}
\begin{center}
\begin{tabular}{ | c || c | c | c | } 

  \hline
  Scenario & \approach & \approach$^-$ & RRT*     \\ 
  \hline
  \hline
  S1 (Fireplace) & $78.0$ & $78.0$ & $85.3\pm3.5$  \\
  \hline
  S2 (Balconies) & $314.8$ & $314.8$ & $372.4\pm47.9$  \\
  \hline
\end{tabular}\label{tab:lengthS1S2}
\end{center}
\end{table}

Table \ref{tab:timeS1S2} presents the results of the comparison between the three planners in terms of total execution time. The RRT * baseline runs for 20 seconds, resulting in a total of 40 seconds for scenario S2, due to the need for two consecutive planning runs. The table illustrates the significant advantage of using the visibility module, which drastically reduces the execution time of the \approach{} strategy, making it at least four times faster in both scenarios, which is of great value in emergency scenarios. In particular, the execution time of \approach{} could be further reduced with a more efficient implementation, such as one in a faster language like C or C++, compared to Python. In view of these results, we conclude that \approach{} outperforms RRT* in both metrics, especially in scenarios with nooks and crannies where the visibility with a catenary is low.

\begin{table}[H]
\caption{Values (in seconds) for the ET metric across the different planners.}
\begin{center}
\begin{tabular}{ | c || c | c | c | } 

  \hline
  Scenario & \approach & \approach$^-$ & RRT*     \\ 
  \hline
  \hline
  S1 (Fireplace) & $0.5$ & $4.4$ & $20$  \\
  \hline
  S2 (Balconies) & $3.2$ & $14.1$ & $40$  \\
  \hline
\end{tabular}\label{tab:timeS1S2}
\end{center}
\end{table}

As an additional test, we ran the RRT* baseline algorithm for scenario S1 with unlimited execution time to determine how long it would take to find a better solution than \approach{}. The baseline approach ultimately found a solution with a path shorter than \approach{} after 18 minutes of computation. This result underscores the substantial impact of addressing the catenary visibility problem for planning tasks with a marsupial robotics system.
\section{Conclusion and Future Work}
\label{sec:conclusions}

An heterogeneous marsupial robotics system uses different types of robots to expand their operation envelope, leveraging the unique strengths of each robot. This paper introduces the planner \approach{}, a novel algorithm for optimal path planning with autonomous tethered marsupial robotic systems. The method computes efficient paths online for specific parameter settings. \approach{} incorporates a visibility module, PVA, to avoid a time-consuming brute-force approach. This module identifies a discrete set of feasible take-off points
after computing maximal visible intervals in the take-off line within a vertical plane. We have demonstrated the effectiveness of \approach{} in both random and realistic scenarios, with experimental results showing that \approach{} outperforms a competitive path planning baseline based on RRT* in terms of both total path length and execution time. An open-source implementation of both \approach{} and the baseline is available online~\footnote{\url{https://github.com/etsi-galgo/maspa}}. 

As future work, we anticipate several potential extensions for \approach{}, which we discuss below:

\begin{itemize}
  \item {\it 3D Visibility Area for Taut Tethers:} Compute the region of feasible \pLvisible{} points in the take-off plane $\pi_{top}$ . Specifically, determine the exact locus of points on the plane from which the target $T$ can be reached with a taut tether of length at most $L$.
  Note that this region is, in general, not connected.

\item{\it Visibility with Catenaries (2D and 3D):} Computing the exact locus of \cLvisible{} points is a challengin geometrical problem. Solving this problem efficiently could enable a more precise discrimination of \noncLvisible{} take-off points compared to the PVA proposed in this work. An exact algorithm for this problem could reduce the execution time and improve the path lengths. Furthermore, computing the exact locus of \cLvisible{} points in the 3D space could potentially solve the problem of finding the optimal solution for SMPP with an efficient and exact algorithm.
\item 
{\it Sequential Targets:}
In scenario S2, we observed the importance of planning sequential paths for multiple target points. In this work, we addressed this problem by planning paths to one target at a time. An interesting extension would be to develop a method that simultaneously optimizes the path for the marsupial to visit all target points. 

\item
{\it Simultaneous Planning:}
Related to the previous point, another open problem is to compute a feasible solution for both the UGV and the UAV considering their simultaneous movement. In this scenario, the UAV would need to visit a set of targets sequentially, while the UGV would have to follow the UAV, ensuring that the paths adhere to the maximum tether length constraint. This problem involves coordinating the movements of both vehicles to optimize the overall path while respecting the length limits of the tether.

\item {\it Field Experimentation:}
Finally, we plan to apply \approach{} to compute trajectories for a real marsupial robotics system prototype. To achieve this, we will integrate the planner with controllers for flight stabilization and trajectory tracking \citep{xu2023event}, ensuring the practical implementation of the proposed algorithm in real-world scenarios.
\end{itemize}

\section*{CRediT Authorship Contribution Statement}



\textbf{Jesús Capitán:} 
Funding acquisition, 
Investigation, 
Writing – original draft.
\textbf{José M. Díaz-Báñez:} 
Project administration, 
Funding acquisition, 
Supervision,  
Writing – review and editing.
\textbf{Miguel A. Pérez-Cutiño:} 
Formal analysis, 
Software, 
Writing – review and editing.
\textbf{Fabio Rodríguez:} 
Formal analysis, 
Visualization, 
Software, 
Writing – original draft.
\textbf{Inmaculada Ventura:} 
Methodology, 
Conceptualization, 
Formal analysis, 
Writing – review and editing.

\section*{Declaration of Competing Interests}
The authors declare that they have no known competing financial interests or personal relationships that could have appeared to influence the work reported in this paper.

\section*{Acknowledgment}

This work is partially supported by the grants PID2020-114154RB-I00 and TED2021-129182B-I00, funded by MCIN/AEI/10.13039/501100011033 and the “European Union
NextGenerationEU/PRTR”.

\bibliography{extra-bib}

\end{document}